\documentclass[10pt,twocolumn,letterpaper]{article}

\usepackage{cvpr}              
\usepackage{footmisc}

\usepackage[table]{xcolor}

\definecolor{cvprblue}{rgb}{0.21,0.49,0.74}
\usepackage[pagebackref,breaklinks,colorlinks,allcolors=cvprblue]{hyperref}

\usepackage{booktabs}  
\usepackage{colortbl}  
\definecolor{lightpink}{RGB}{233, 23, 115}
\definecolor{first}{RGB}{220, 245, 220}   
\definecolor{second}{RGB}{255, 255, 200}  
\definecolor{third}{RGB}{200, 230, 255}   
\definecolor{lightblue}{RGB}{100, 149, 237} 
\definecolor{standardblue}{RGB}{0, 112, 192}
\definecolor{lightgreen}{RGB}{144,238,144}

\def\paperID{4601} 
\def\confName{CVPR}
\def\confYear{2026}

\title{
AdaPower: Specializing World Foundation Models for Predictive Manipulation 
}

\author{Yuhang Huang\footnotemark[1]\\
National University of\\ Defense Technology
\and
Shilong Zou\footnotemark[1]\\
National University of\\ Defense Technology
\and
Jiazhao Zhang\\
Peking University
\and
Xinwang Liu\\
National University of\\ Defense Technology
\and
Ruizhen Hu\footnotemark[2]\\
Shenzhen University
\and
Kai Xu\footnotemark[2]\\
National University of\\ Defense Technology
}

\begin{document}
\maketitle
\footnotetext[1]{Equal contributions.}
\footnotetext[2]{Co-corresponding authors.}
\begin{abstract}
World Foundation Models (WFMs) offer remarkable visual dynamics simulation capabilities, yet their application to precise robotic control remains limited by the gap between generative realism and control-oriented precision. While existing approaches use WFMs as synthetic data generators, they suffer from high computational costs and underutilization of pre-trained VLA policies. We introduce \textbf{AdaPower} (\textbf{Ada}pt and Em\textbf{power}), a lightweight adaptation framework that transforms general-purpose WFMs into specialist world models through two novel components: Temporal-Spatial Test-Time Training (TS-TTT) for inference-time adaptation and Memory Persistence (MP) for long-horizon consistency. Integrated within a Model Predictive Control framework, our adapted world model empowers pre-trained VLAs, achieving over 41\% improvement in task success rates on LIBERO benchmarks without policy retraining, while preserving computational efficiency and generalist capabilities.

\end{abstract}

\section{Introduction}

The emergence of World Foundation Models (WFMs)~\cite{agarwal2025cosmos, cogvideox, wan2025wan, chi2025wow, yang2023unisim} represents a paradigm shift in visual dynamics simulation, with models trained on internet-scale video datasets demonstrating unprecedented capabilities in capturing physical and semantic regularities of the world. These models exhibit exceptional generalization across diverse environments through their ability to predict visual scene evolution conditioned on multimodal prompts. However, a critical challenge arises when deploying these powerful generative models in robotic manipulation domains that demand continuous, fine-grained environment interactions. The fundamental disconnect lies in the contrasting requirements: while WFMs excel at producing \textit{visually plausible} future predictions, robotic control necessitates \textit{action-conditioned, precisely executable} dynamic forecasts. This dichotomy between generative capability and control-oriented reliability presents a significant barrier to direct WFM application in robotics.

\begin{figure}[t]
    \centering
    \includegraphics[width=01\linewidth]{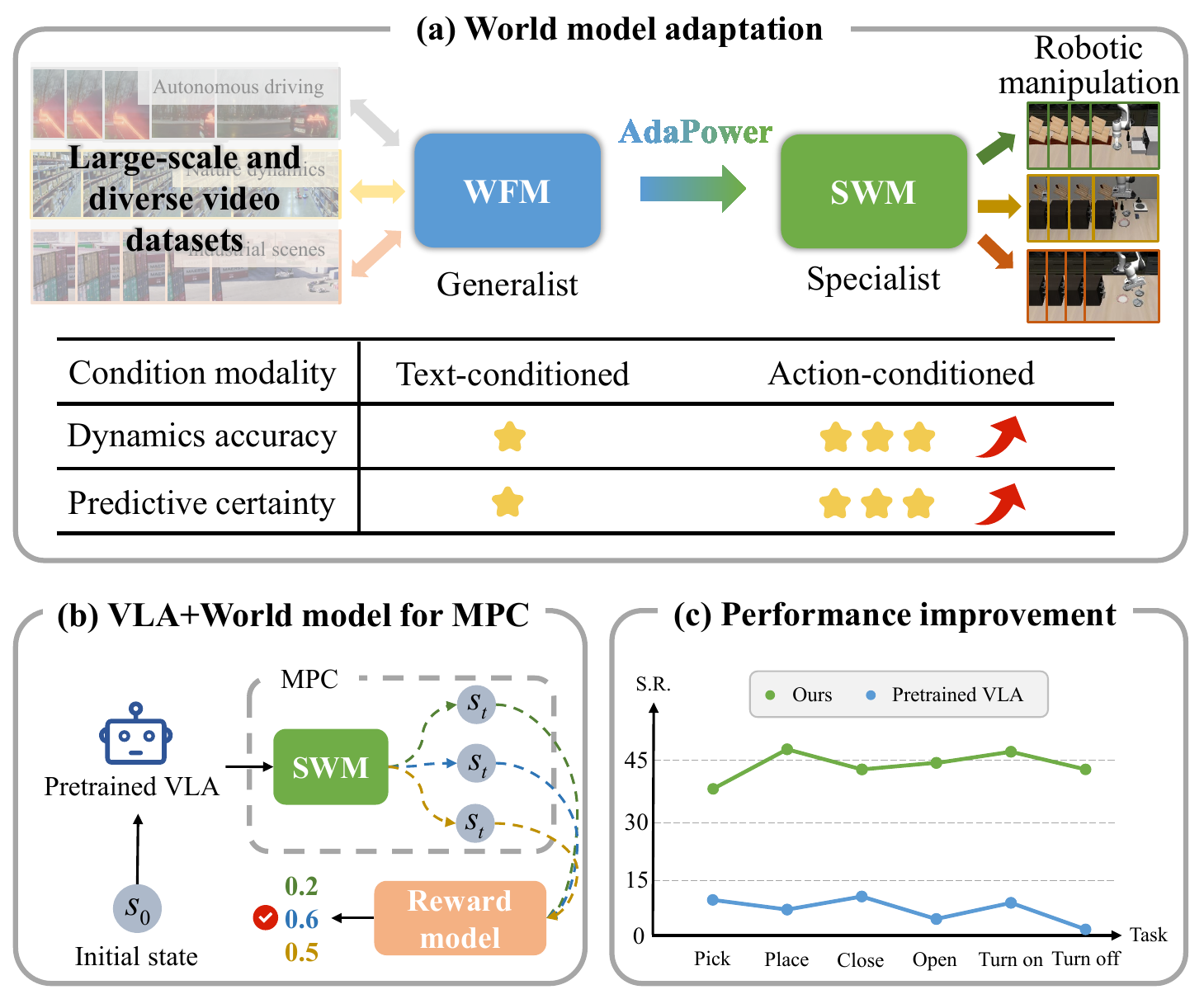}
    \caption{Illustration of our method. (a) We propose AdaPower, an efficient adaptation framework that transforms a generalist world foundation model (\textbf{WFM}) into a specialist world model (\textbf{SWM}), improving both dynamic accuracy and certainty. (b) The specialist world model can serve as a dynamic model for model predictive control, (c) which significantly improves the zero-shot generalization of the  pretrained VLA model.}
    \label{fig:teaser}
\end{figure}

Current state-of-the-art approaches, exemplified by DreamGen~\cite{jang2025dreamgen}, adopt a synthetic data generation paradigm to bridge this gap. The conventional pipeline involves fine-tuning WFMs on limited target-robot data, generating synthetic videos using the adapted model, extracting pseudo-action sequences through inverse dynamics models, and training visuomotor policies from the resulting neural trajectories. While demonstrating promising results, this paradigm suffers from three fundamental limitations: 1) the \emph{prohibitive computational costs} associated with generating massive video datasets and training new policies; 2) \emph{lengthy adaptation cycles} requiring complete pipeline repetition for each new task or environment; and 3) \emph{underutilization of powerful pre-trained policies} by bypassing the potential to enhance existing generalist VLA~\cite{li2024cogact, black2024pi_0, nvidia2025gr00t, bu2025univla} models in favor of training new specialist policies.

We propose \textbf{AdaPower} (\textbf{Ada}pt and Em\textbf{power}), a framework that diverges from the data-generation paradigm by focusing on efficient model adaptation to empower existing policies (Fig.~\ref{fig:teaser}). Our core insight is that through lightweight parameter optimization, we can transform general-purpose WFMs into specialist world models that directly improve the zero-shot capabilities of pre-trained VLAs. This approach offers compelling advantages: 1) \emph{computational efficiency} through minimal parameter adaptation; 2) \emph{preservation of generalist knowledge} inherent in pre-trained VLAs; and 3) \emph{true zero-shot improvement} by enhancing performance on novel environment without retraining.

This adaptation approach nevertheless faces two fundamental challenges derived from using WFMs in closed-loop control settings: 1) \emph{test-time distribution shift}, where models encounter novel initial states and environmental configurations unseen during adaptation, causing predictions to rapidly degrade without mechanisms for self-supervised optimization; and 2) \emph{long-horizon inconsistency}, where minor errors in multi-step autoregressive rollouts accumulate, leading to physically implausible predictions that undermine planning stability.

Our AdaPower framework addresses these challenges through two innovative components. The \emph{Temporal-Spatial Test-Time Training (TS-TTT)} module enables self-supervised optimization across spatial and temporal dimensions during inference, allowing adaptation to novel test-time distributions. The \emph{Memory Persistence (MP)} module maintains historical context through cross-attention, ensuring long-horizon consistency by preventing error accumulation across prediction steps. The adapted specialist model then integrates into a Model Predictive Control framework where it functions as a dynamics simulator alongside a pre-trained VLA policy, creating a cooperative system where the VLA serves as a high-level planner proposing candidate action sequences, while the adapted world model performs precise rollouts to evaluate physical feasibility.

Extensive evaluation across diverse simulated and real-world manipulation tasks demonstrates that AdaPower achieves over 41\% improvement in success rates while maintaining computational efficiency and enhancing the generalization of pre-trained VLA policies. Our work establishes a new paradigm for leveraging world foundation models in robotic control, effectively bridging the gap between general world knowledge and specialist robotic requirements.

In summary, our main contributions include:
\begin{itemize}
\item We introduce AdaPower, a novel framework that efficiently adapts WFMs into specialist world models, establishing a new paradigm for leveraging internet-scale video priors in robotic manipulation tasks.
\item We design Temporal-Spatial Test-Time Training (TS-TTT) and Memory Persistence (MP) modules that effectively address the core challenges of test-time adaptation and long-horizon coherence.
\item We develop a cooperative MPC system that empowers pre-trained VLAs with strong zero-shot generalization, demonstrating significant performance gains across diverse settings and bridging the gap between generalist world knowledge and specialist robotic control.
\end{itemize}    
\begin{figure*}[th]
    \centering
    \includegraphics[width=\textwidth]{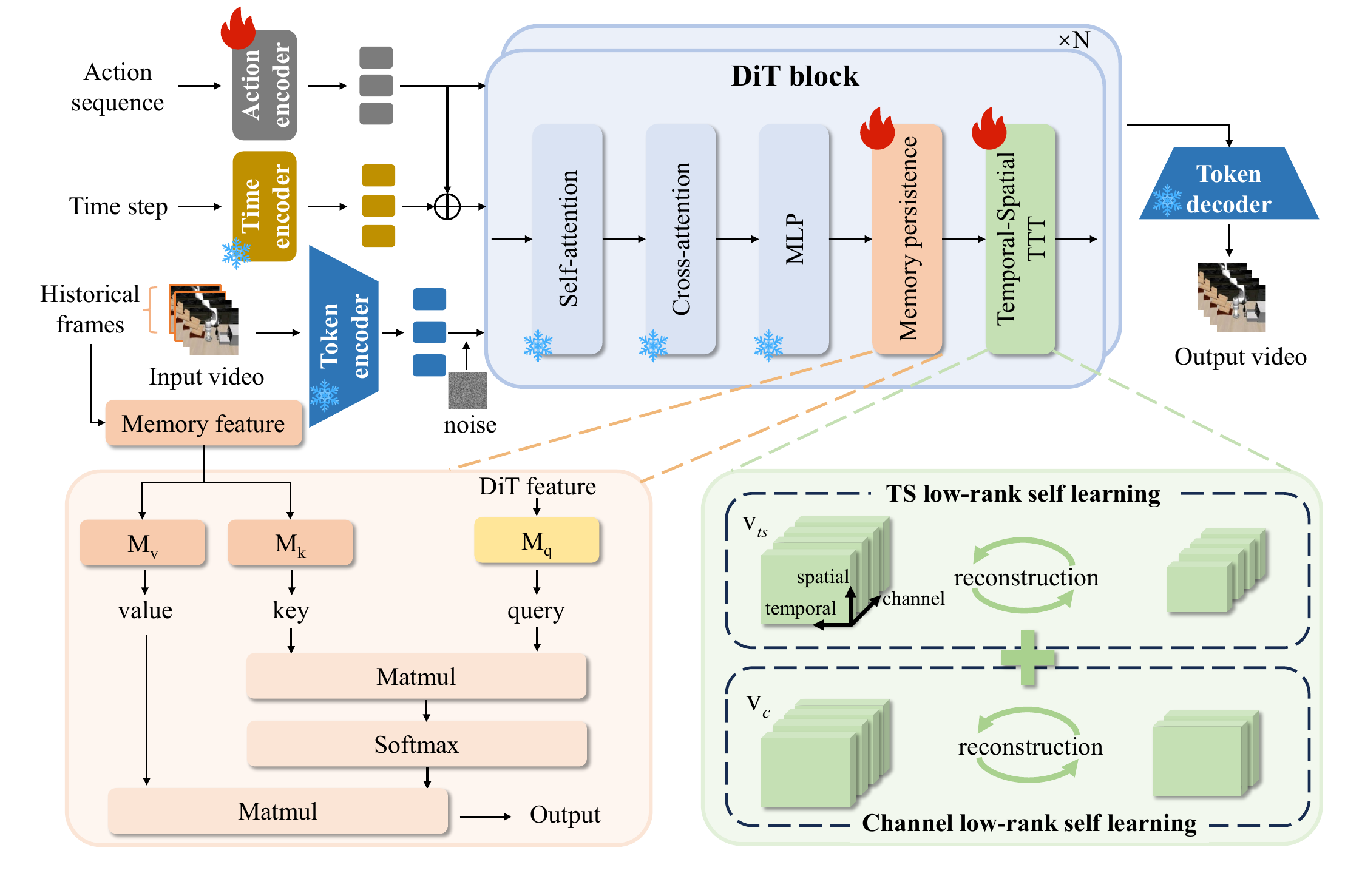}
    \caption{Overall architecture. We introduce two key components based on the diffusion transformer architecture of Cosmos-Predict-2B \cite{agarwal2025cosmos}. The temporal–spatial TTT enables the model to optimize partial parameters through a self-supervised loss across the temporal–spatial dimension during inference, thereby improving data efficiency and generalizability. The memory persistence module integrates historical information into a memory feature and allows the DiT feature to interact with it to preserve historical context.}
    \label{fig:pipeline}
\end{figure*}
\section{Related Work}
\textbf{World foundation models}.
Recent world foundation models (WFMs) trained on Internet-scale video data advance generative dynamics modeling and interactive simulation. Diffusion-based video generators~\cite{ho2022video, zhang2025waver, gao2024vista, wu2024ivideogpt} and scalable transformer backbones~\cite{peebles2023scalable} provide strong visual dynamics priors and long-horizon coherence.
Cosmos~\cite{agarwal2025cosmos} introduces a world foundation model platform designed to support the development of customized world models for Physical AI; WAN~\cite{wan2025wan} proposes a comprehensive suite of open-source video foundation models based on diffusion transformers, achieving state-of-the-art video generation performance through large-scale pre-training and efficient architectures.
In interactive settings, WFMs increasingly accept action or multimodal prompts, e.g., GAIA-1~\cite{hu2023gaia} for autonomous driving and DeepMind’s GENIE~\cite{bruce2024genie} for action-conditioned virtual environments. For robotics, Genie Envisioner~\cite{liao2025genie} builds a unified world foundation platform that bridges world modeling and policy learning through a video-generative diffusion framework, offering an integrated solution for instruction-driven robotic manipulation. Together, these lines motivate adapting generalist WFMs into action-conditioned, control-ready specialists for manipulation while preserving broad priors.

\textbf{World models for model predictive control}. World models have become central to model-based planning, enabling model predictive control (MPC) through learned dynamics. TD-MPC \cite{hansen2022tdmpc} introduces a unified framework that jointly learns latent dynamics and performs gradient-based trajectory optimization via temporal-difference value expansion, achieving strong data efficiency and accuracy. Its successor, TD-MPC2 \cite{hansen2022tdmpc}, improves stability with robust value gradients and policy-consistency regularization, further enhancing long-horizon control. Beyond TD-MPC, MBOP \cite{argenson2021mbop} learns an ensemble of dynamics models from offline data and uses them within an MPC planner constrained by a behavior-cloned policy prior. MoPA \cite{nagabandi2020mopa} integrates a learned dynamics model into a differentiable MPC controller for gradient-based action optimization. More recent works such as MWM \cite{rybkin2023mwm}, GDM \cite{lambert2022gdm}, and V-JEPA2 \cite{assran2025v} develop modular, end-to-end differentiable world models for robust planning. Our AdaPower builds upon this paradigm, enabling world foundation models to serve as high-fidelity, action-conditioned dynamics predictors for MPC.

\textbf{Adapter for large models}.
Adapter for large models enables specialization without full model updates. In vision, AdaptFormer~\cite{chen2022adaptformer} augments ViTs with lightweight MLP adapters; Visual Prompt Tuning~\cite{jia2022visual} learns input-space prompts, both retaining frozen backbones while matching or surpassing full fine-tuning under budget limits. In diffusion models, ControlNet~\cite{zhang2023adding} introduces conditional control branches that preserve pretrained knowledge yet add spatial constraints, and T2I-Adapter similarly aligns external control signals with frozen text-to-image models. VACE~\cite{vace} employs unified adapter layers that inject motion and temporal cues into pretrained text-to-video diffusion models. Beyond the above, LoRA’s low-rank updates have become a general parameter-efficient fine-tuning primitive used across image and video diffusion models. These techniques suggest that adapters—attached to diffusion transformer blocks—can inject action conditioning, temporal memory, and task signals into WFMs with minimal compute and reduced overfitting risk, aligning with our MP-TTT design.

\textbf{Test-time training}.
Different from general adapters that update models only during the training stage, Test-Time Training (TTT)~\cite{sun2024learning} also updates models on unlabeled test inputs to counter distribution shifts with minimal supervision at test time. The seminal TTT formulates self-supervised objectives at inference; TENT~\cite{wang2021tent} minimizes predictive entropy online. For non-stationary or prolonged deployment, CoTTA~\cite{wang2022continual} adds augmentation-averaged predictions and stochastic weight restoration to curb error accumulation and forgetting. EATA~\cite{niu2022efficient} further improves efficiency and stability via active sample selection and Fisher-regularized updates, with follow-ups exploring uncertainty-calibrated variants. TTT-MLP~\cite{dalal2025one} explores enhancing long-form video generation by introducing TTT layers into pre-trained Transformers, enabling the generation of coherent one-minute videos from text storyboards. Our Temporal-Spatial TTT module builds on these principles but targets video dynamics and action-conditioned rollouts, optimizing over spatiotemporal features to adapt world-model predictions during MPC planning.

\section{Method}
\label{sec:method}
We present \textbf{AdaPower}, an efficient framework to adapt a world foundation model (WFM) into a specialist world model (SWM) and thus empower robotic manipulations. 
Section~\ref{sec:architeture} presents the overall architecture of the proposed system and illustrates how the adapter, with two key components: temporal-spatial test-time training (TS-TTT) module and the memory persistence (MP) module, is integrated into the WFM. 
Section~\ref{sec:TTT} introduces the TS-TTT module, extending conventional TTT to exploit the intrinsic low-rank structure of video data across both spatial and temporal dimensions. 
Section~\ref{sec:MP} details the MP module, which maintains historical information through cross-attention to ensure temporal consistency.  
Finally, Section~\ref{sec:MPC} describes the deployment strategy with Model Predictive Control (MPC), where our specialist world model collaborates with a pre-trained Vision-Language-Action (VLA) model to enhance zero-shot generalization in robotic manipulation tasks.

\subsection{Overall Architecture}
\label{sec:architeture}
Fig.~\ref{fig:pipeline} provides an overview of our \textbf{AdaPower} architecture, where the adapter is inserted into the backbone of a Diffusion Transformer (DiT)-based world model. We utilize Cosmos-Predict2-2B \cite{agarwal2025cosmos} as our base world foundation model, as it has been trained on large-scale datasets to learn general physical priors.

The adapter consists of two key modules: the memory persistence (MP) module and the temporal-spatial test-time training (TS-TTT) module, responsible for enhancing long-term predictive consistency and generalization ability, respectively. Moreover, to enable action-conditioned prediction, we replace the original text encoder of the WFM with an action encoder, which consists of multiple perception layers. 
This modification transforms the specialist world model into an action-aware predictive model suitable for manipulation tasks.

During adaptation, only the newly introduced layers, including the MP module, the TS-TTT modual, and the action encoder, are trained, while the pre-trained WFM parameters remain frozen.
This ensures high computational efficiency and maintains the general understanding of physical dynamics while specializing in manipulation-specific reasoning.

\subsection{Temporal-Spatial Test-Time Training}
\label{sec:TTT}
Test-Time Training (TTT)~\cite{sun2024learning} is a self-optimized adaptation technique that updates a subset of model parameters during inference by minimizing a self-supervised loss.
Typical TTT designs a reconstruction loss that projects the input into a low-rank representation and learns to reconstruct the input from this low-rank variable.
However, conventional TTT is designed for sequence modeling and learns only the \emph {channel-wise} low-rank prior using 2D feature ($T\times D$), which cannot directly used to fully caputre the low-rank priors in the \emph{spatial} and \emph{temporal} dimensions of 4D video features ($T\times H\times W\times D$) when dealing with world models.

Our key observation is that video feature exhibits various low-rank priors: (a) \emph{spatial low-rank}—adjacent pixels often share similar colors and textures within the same frame; and (b) \emph{temporal low-rank}—most background content remains static or changes gradually between consecutive frames.
Therefore, we propose Temporal–Spatial Test-Time Training (TS-TTT) for world modeling, which extends conventional TTT with temporal–spatial low-rank self-learning. TS-TTT enables the model to jointly learn low-rank priors across temporal–spatial and channel-wise dimensions, thereby amplifying the self-supervised learning signal and improving generalization to unseen environments. 


In more details, TS-TTT consists of two branches of self-learning optimization: one for the temporal–spatial dimensionand another for the channel dimension.
Specifically, given the video input feature $\mathbf{v}\in\mathbb{R}^{T\times H\times W\times D}$, where $T$ denotes the temporal dimension, $H\times W$ represent the spatial dimensions, and $D$ is the channel dimension, we first flatten $\mathbf{v}$ into a 2D sequence $\mathbf{v}_{ts}=\{v_{ts}^{i}\}_{i=1}^{D}$, where $\mathbf{v}_{ts}\in\mathbb{R}^{D\times THW }$, and then perform TTT along the temporal-spatial dimension. 
Note that we posit that the temporal and spatial dimensions of a video are correlated; that is, objects exhibit similar appearances and geometries across spatial and temporal movements. Therefore, we fuse the temporal and spatial dimensions for self-learning (additional ablation results are provided in Sec.~\ref{sec:ablation}).

In practice, we define the TS-TTT layer using a neural network $f$ parameterized by $W$, which map any input token $v_{ts}$ to an output token $z_{ts}$: 
\begin{equation}
    z_{ts} = f(W;v_{ts}).
    \label{eq1}
\end{equation}
The key idea is to optimize $W$ iteratively based on the historical context $v_{ts}^{1},..., v_{ts}^{i}$ through a self-supervised loss $\ell$:
\begin{equation}
    W^i = W^{i-1} - \eta\nabla\ell(W^{i-1}; v_{ts}^{i}),
    \label{eq2}
\end{equation}
where $\eta$ is the learning rate. For the format of the self-supervised loss $\ell$, a general selection is reconstructing $v_{ts}^{i}$ itself. To make the learning problem nontrivial, one can first process $v_{ts}^{i}$ into a corrupted variable $\hat{v}_{ts}^{i}$, thus we can express $\ell$ as:
\begin{equation}
    \ell(W; v_{ts}^{i}) = \Vert f(W; \hat{v}_{ts}^{i})-v_{ts}^{i} \Vert^2.
    \label{eq3}
\end{equation}
Similar to denoising autoencoders~\cite{vincent2008extracting}, the function $f$ must discover correlations between the temporal-spatial dimensions of $v_{ts}^{i}$ to reconstruct it from partial information. 
In practice, $\hat{v}_i$ in Eq.~\ref{eq3} is obtained via a low-rank projection $\hat{v}_i=\Theta_{K}v_{ts}^{i}$, where $\Theta_K$ is a learnable matrix. Moreover, considering that not all the information in $v_{ts}^{i}$ is worth remembering, so the reconstruction label can also be another projection $\Theta_{V}v_{ts}^{i}$ instead of $v_{ts}^{i}$ itself. In summary, the self-supervised loss  $\ell$ is:
\begin{equation}
    \ell(W; v_{ts}^{i}) = \Vert f(W; \Theta_{K}v_{ts}^{i})-\Theta_{V}v_{ts}^{i} \Vert^2.
    \label{eq4}
\end{equation}
Finally, because $\Theta_{K}v_{ts}^{i}$ has a lower dimension than $v_{ts}^{i}$, we can no longer utilize the output rule defined in Eq.~\ref{eq1}. Consequently, an additional projection $\Theta_{Q}v_{ts}^{i}$ is introduced to modify the output rule as follows:
\begin{equation}
    z_{ts}^{i} = f(W^i; \Theta_{Q}v_{ts}^{i}).
    \label{eq5}
\end{equation} 
Note that only $W$ is optimized during reference, while $\Theta_{Q},\Theta_{K},\Theta_{V}$ are all optimized together with the general parameters of the base network during training.
Combining Eq.~\ref{eq2}, ~\ref{eq4} and ~\ref{eq5}, the temporal-spatial branch produces the output sequence $\mathbf{z}_{ts}=\{z_{ts}^{i}\}_{i=1}^{D}$. 

For the channel-wise branch, 
we reshape $\mathbf{v}$ into $\mathbf{v}_{c}=\{v_{c}^{t}\}_{t=1}^{THW}$, where $\mathbf{v}_{c}\in\mathbb{R}^{THW\times D }$, and perform low-rank projection and reconstruction following Eq.~\ref{eq2}, \ref{eq4}, and \ref{eq5} with its own set of parameters, yielding the output $\mathbf{z}_{c}=\{z_{c}^{t}\}_{t=1}^{THW}$. Finally, we reshape $\mathbf{z}_{ts}$ and $\mathbf{z}_{c}$ back to the same shape as $\mathbf{v}$, and design the final output $\mathbf{v}^{o}$ with a residual connection:
\begin{equation}
    \mathbf{v}^{o} = \mathbf{v} + \mathbf{z}_{ts} + \mathbf{z}_{c},
\end{equation}
This design enables our method to conduct self-supervised optimization across temporal, spatial and channel-wise dimensions.

\subsection{Memory Persistence Module}
\label{sec:MP}
The ability to perform long-horizon prediction is crucial for predictive manipulation, however, forecasts produced by world models often suffer from error accumulation and knowledge forgetting, where earlier observations gradually lose influence as rollouts proceed.  
To alleviate this issue, we further introduce the Memory Persistence (MP) module.

As shown in Fig.~\ref{fig:pipeline}, at each step, the MP module retrieves historical memory features from prior frames and fuses them with current DiT block features via cross-attention. This enables coherent scene representations over time, ensuring temporal consistency and preserving critical object relationships. By integrating historical context directly into each DiT layer, MP stabilizes multi-step rollouts and reduces drift common in autoregressive prediction.

 More specifically, 
since the environmental dynamics are primarily driven by interactions between the robotic agent and surrounding objects, we aim to extract object-centric information as the memory feature. The recent visual foundation model DINOv2~\cite{dinov2} has demonstrated remarkable capability in grounding object-centric features; therefore, we employ DINOv2 as an encoder to extract the memory feature. 
Thus, the historical frames $\mathbf{H}\in\mathbb{R}^{L\times H\times W\times3}$ (where $L$ denotes the number of historical frames) are fed into the DINOv2 model, yielding dense patch tokens $\mathbf{m}\in\mathbb{R}^{L\times P\times D}$ (where $P$ represents the number of patch tokens). The memory feature is obtained by reshaping $\mathbf{m}$ into a 2D sequence with dimensions $LP\times D$, enabling the subsequent attention operation. We then apply cross-attention to allow the DiT feature to integrate historical context, i.e., the memory feature produces the key and value representations, while the DiT feature generates the query representation, following the standard attention formulation~\cite{vaswani2017attention}.

\begin{figure}[t]
    \centering
    \includegraphics[width=1\linewidth]{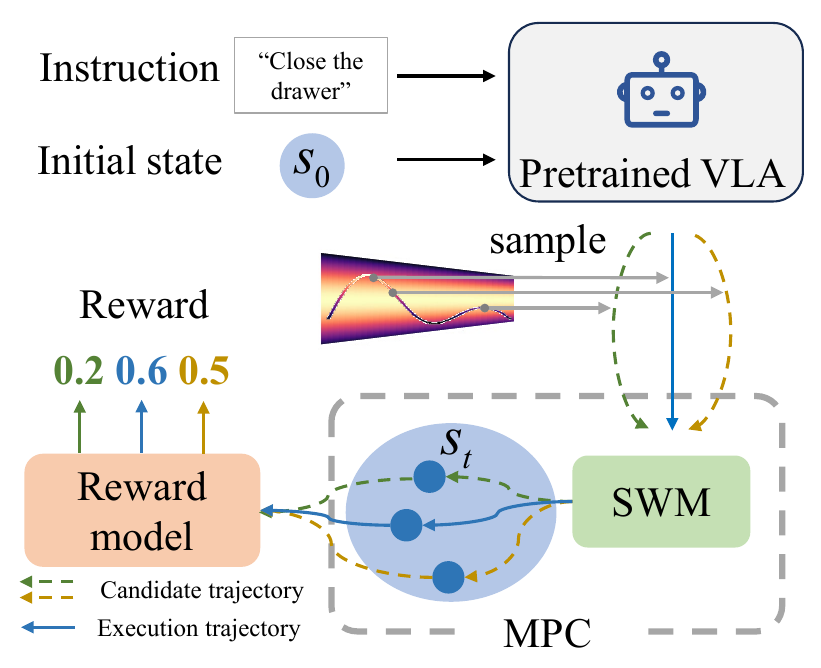}
    \caption{Framework of MPC. The pretrained VLA generates candidate actions from an initial state and task instruction. Then, a specialist world model simulates these actions into trajectories, which a reward model evaluates to select the optimal sequence for execution.}
    \label{fig:mpc}
\end{figure}

\begin{table*}[th]
    \centering
    \caption{Performance comparisons on 10 LIBERO tasks. The first row presents the results of the pretrained VLA, while the other rows show the results of MPC with the specialist world models powered by different adapters.}
    \label{tab:1}
    \renewcommand{\arraystretch}{1.0}
    \setlength{\tabcolsep}{5pt}
    \begin{tabular}{l|ccccccccccc}
\toprule
Method                                                      & Task1 & Task2 & Task3 & Task4 & Task5 & Task6 & Task7 & Task8 & Task9 & Task10 & Avg.SR \\ \hline\hline
\begin{tabular}[c]{@{}l@{}}Pretrained VLA\end{tabular} & 0     & 0     & 0     & 0     & 0     & 0     & 5     & 0     & 0     & 0      & 0.5    \\ \hline
+ VACE                                                 & 15    & 20    & 15    & 15    & 15    & 15    & 30    & 25    & 25    & 25     & 20.5   \\
+ LoRA                                                 & 20    & 20    & 25    & 10    & 20    & 25    & 35    & 25    & 30    & 25     & 23.5   \\
+ SFT                                                 & 20    & 30    & 40    & 10    & 25    & 35    & 50    & 35    & 40    & 40     & 32.5   \\
\rowcolor{gray!20}
+ AdaPower (Ours)                                                        & 25    & 40    & 50    & 20    & 35    & 40    & 60    & 50    & 55    & 40     & 41.5   \\ \bottomrule
\end{tabular}
\end{table*}

\begin{figure*}[t]
    \centering
    \includegraphics[width=\linewidth]{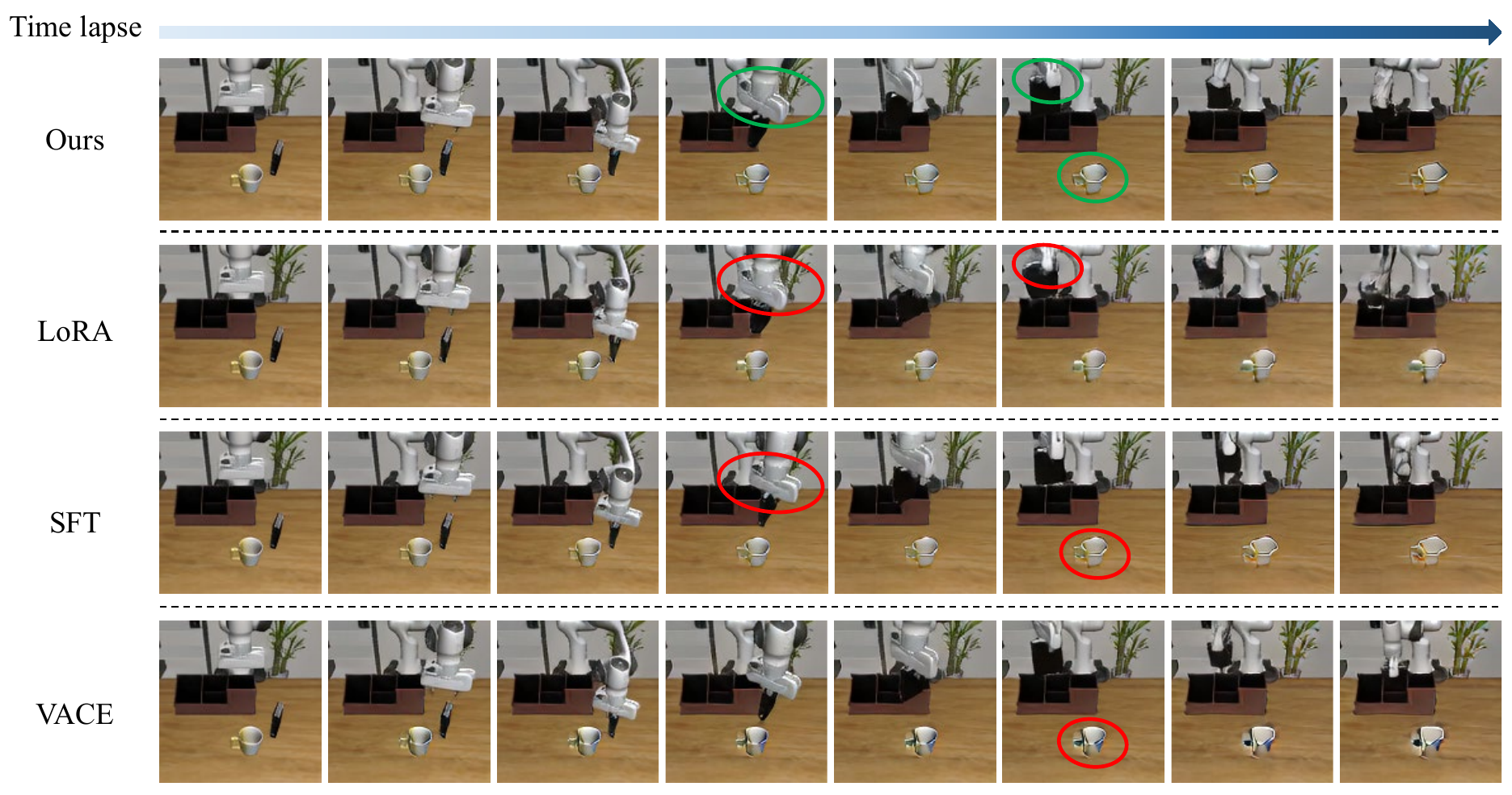}
    \caption{Qualitative comparisons of rollouts generated by different specialist world models. We evaluate these models on unseen tasks from LIBERO-LONG to assess their long-horizon prediction performance. We highlight the improvements of our method using circles.}
    \label{fig:comp}
\end{figure*}

\subsection{Deployment with Model Predictive Control}
\label{sec:MPC}

After adaptation, the specialist world model is integrated into a Model Predictive Control (MPC) framework for robotic manipulation. As shown in Fig.~\ref{fig:mpc}, our system employs a pre-trained VLA model as a high-level planner that generates candidate action sequences from task instructions. The adapted world model acts as a dynamics predictor, performing rollouts for each candidate to forecast future states and identify the optimal control trajectory.

Specifically, given an initial state and task instruction, the VLA produces multiple candidate action sequences. To enhance exploration diversity, we augment these with noisy sequences sampled from a low-variance Gaussian distribution. The specialist world model then simulates these sequences to generate imagined trajectories. For evaluation, we use a reward model based on the ReWind approach~\cite{zhang2025rewind}, which estimates task completion progress from task instructions and historical observations. The action sequence with the highest predicted reward is selected for execution.
This collaborative architecture combines the VLA's general reasoning with the specialist world model's accurate physical simulation. Notably, the VLA operates in a zero-shot manner without fine-tuning, while the world model's improved generalization enables adaptation to unseen scenarios. 

\begin{table*}[th]
    \centering
    \caption{Ablation studies of module effectiveness. }
    \label{tab:ablation}
    \renewcommand{\arraystretch}{1.0}
    \setlength{\tabcolsep}{5pt}
    \begin{tabular}{cc|ccccccccccc}
\toprule
TS-TTT &  MP     & Task1 & Task2 & Task3 & Task4 & Task5 & Task6 & Task7 & Task8 & Task9 & Task10 & Avg.SR \\ \hline\hline
\checkmark      & $\times$                                       & 20    & 40    & 50    & 20    & 30    & 40    & 60    & 45    & 55    & 40     & 40.0   \\
$\times$      & \checkmark                                      & 20    & 40    & 40    & 15    & 25    & 40    & 55    & 50    & 55    & 40     & 38.0   \\
\rowcolor{gray!20}
\checkmark & \checkmark                                          & 25    & 40    & 50    & 20    & 35    & 40    & 60    & 50    & 55    & 40     & 41.5  \\ \bottomrule
\end{tabular}
\end{table*}

\section{Experiment}
\subsection{Experimental Setup}
\textbf{Model architecture}. To achieve an optimal balance between efficiency and adaptation capability, we do not insert the proposed modules into every block of the Diffusion Transformer (DiT) backbone. Instead, both the TS-TTT and MP modules are inserted every seven blocks, allowing them to enhance temporal reasoning without excessively increasing parameter count or computational cost.
This design adds approximately 150M new parameters, which is less than 10\% of the base model size.
To maintain stability during early training, the outputs of the newly added layers are zero-initialized.
This initialization ensures that, at the start, the added modules do not perturb the pretrained representations, gradually learning to contribute as optimization progresses.
This simple yet effective technique was crucial for avoiding divergence during the first few thousand iterations.

\textbf{Training configuration}. For simulation experiments, we use 2,000 trajectories sampled from the LIBERO-90 \cite{liu2023libero} dataset—a large-scale robotic manipulation dataset covering a diverse range of objects, motions, and contact-rich interactions.
The model is fine-tuned for 10,000 iterations using AdamW optimizer.
We set the learning rate to 1e-3 for TS-TTT parameters and 1e-4 for all other trainable components. A linear learning rate scheduler with warm-up is applied for stable convergence.

\begin{table}[t]
    \centering
    \caption{Different variants of TS-TTT. ``T+S+C'' means that three separate branches are used to process the temporal (T), spatial (S), and channel (C) dimensions independently. ``T+SC'' means that the spatial and channel dimensions are fused into a single branch. ``TSC'' means that a single branch processes all three dimensions.}
    \label{tab:TS-TTT}
    \renewcommand{\arraystretch}{1.2}
    \setlength{\tabcolsep}{5pt}
    \begin{tabular}{l|cccc}
    \toprule
    Method & T+S+C & T+SC & TSC & TS+C (ours) \\ \hline\hline
    Test MSE &  0.0613  & 0.1244  &  NaN  &  0.0205   \\ \hline
    Avg.SR &  35.0  &  15.0  &  0  &  41.5     \\ \bottomrule
    \end{tabular}
\end{table}

\textbf{Evaluation protocol}. At evaluation time, we test the adapted specialist world model on ten unseen manipulation tasks.
These tasks are excluded from the fine-tuning set to evaluate zero-shot generalization of the adapted model.
Each task is executed 20 times, and the average task success rate is reported as the primary metric.
All methods are evaluated under the same Vision-Language-Action (VLA) policy and MPC planner to ensure fairness.
Detailed task definitions and hyper-parameters are provided in Supplementary Material.



\subsection{Comparisons with other Adapters}
We use CogACT~\cite{li2024cogact} as the base VLA model to conduct the simulation experiments. Tab.~\ref{tab:1} summarizes the quantitative results comparing AdaPower with existing fine-tuning approaches, including VACE~\cite{vace}, LoRA~\cite{hu2022lora}, and supervised fine-tuning (SFT). 
Our method achieves the highest performance across all ten tasks, improving the pre-trained VLA’s overall success rate by 41\%.

VACE and LoRA are two parameter-efficient adapters that keep the pretrained weights fixed and train only the newly added layers. Compared with these methods, our approach incorporates a unique memory persistence mechanism and a test-time optimization strategy, achieving more accurate predictions on unseen environment. The SFT method exhibits reasonable performance but requires substantial computational resources. In contrast, our method is more efficient and enables the world model to perform self-learning on unseen tasks, leading to better results.
Fig.~\ref{fig:comp} illustrates long-horizon rollouts of unseen tasks across different methods. 
The compared methods tend to accumulate small errors that eventually lead to blur objects or physical inconsistency.
In contrast, our method maintains stable object trajectories over extended horizons, correctly modeling contact dynamics and force propagation. These results highlight the adapter’s ability to preserve temporal coherence while enhancing generalizable prediction.

\begin{figure}[t]
    \centering
    \includegraphics[width=01\linewidth]{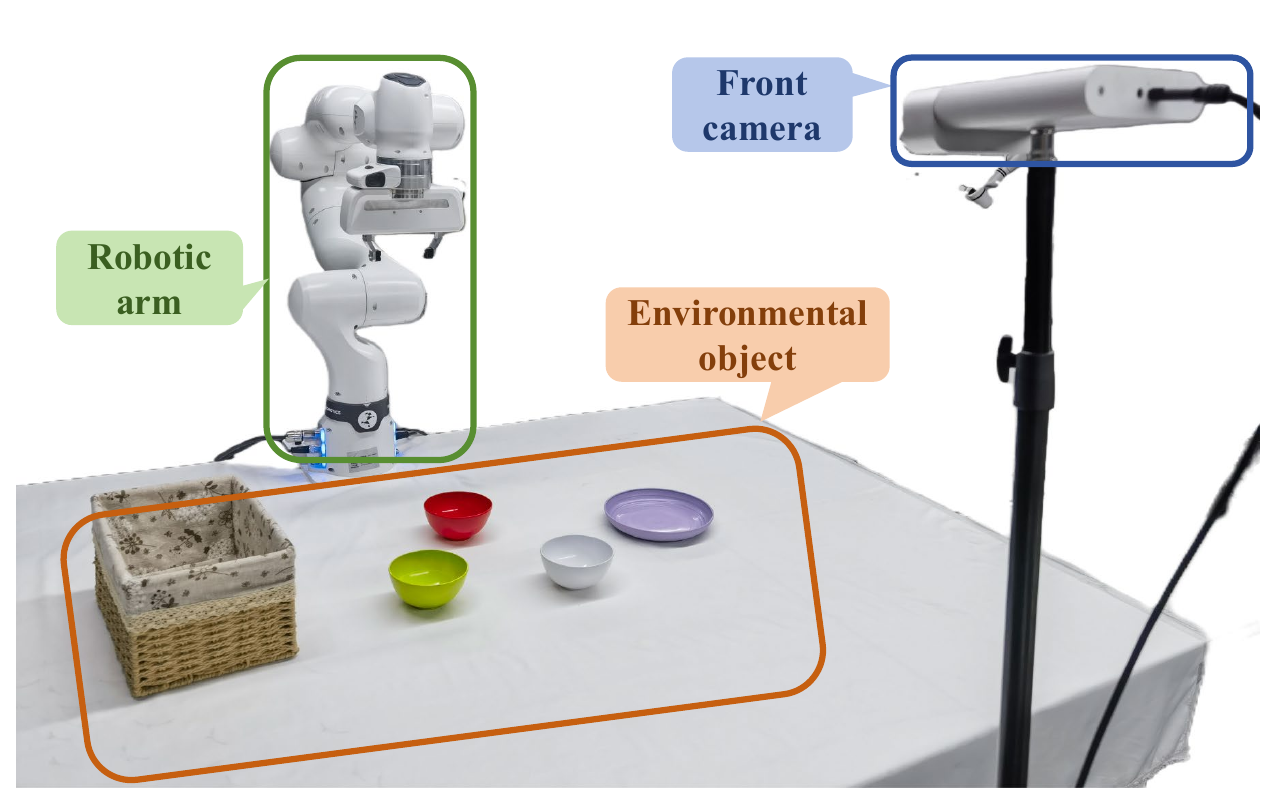}
    \caption{Real-world setup. We utilize a 7-degree-of-freedom robotic arm to perform real-world manipulation tasks, including placing, grasping, and stacking.}
    \label{fig:setup}
\end{figure}

\begin{table*}[th]
    \centering
    \caption{Evaluation on the deployment with different initial VLA models.}
    \label{tab:4}
    \renewcommand{\arraystretch}{1.0}
    \setlength{\tabcolsep}{5pt}
    \begin{tabular}{l|ccccccccccc}
\toprule
Method                                                      & Task1 & Task2 & Task3 & Task4 & Task5 & Task6 & Task7 & Task8 & Task9 & Task10 & Avg.SR \\ \hline\hline
CogACT & 0     & 0     & 0     & 0     & 0     & 0     & 5     & 0     & 0     & 0      & 0.5    \\
+ AdaPower                                                       & 25    & 40    & 50    & 20    & 35    & 40    & 60    & 50    & 55    & 40     & 41.5   \\ \hline
$\pi_0$ & 5     & 15     & 20     & 0     & 25     & 20     & 45     & 20     & 40     & 20      & 21.0   \\
+ AdaPower                                                       & 20    & 45    & 50    & 25    & 45    & 50    & 70    & 50    & 75    & 50     & 48.0  \\ 
\bottomrule
\end{tabular}\vspace{-10pt}
\end{table*}

\begin{figure*}[th]
    \centering
    \includegraphics[width=\linewidth]{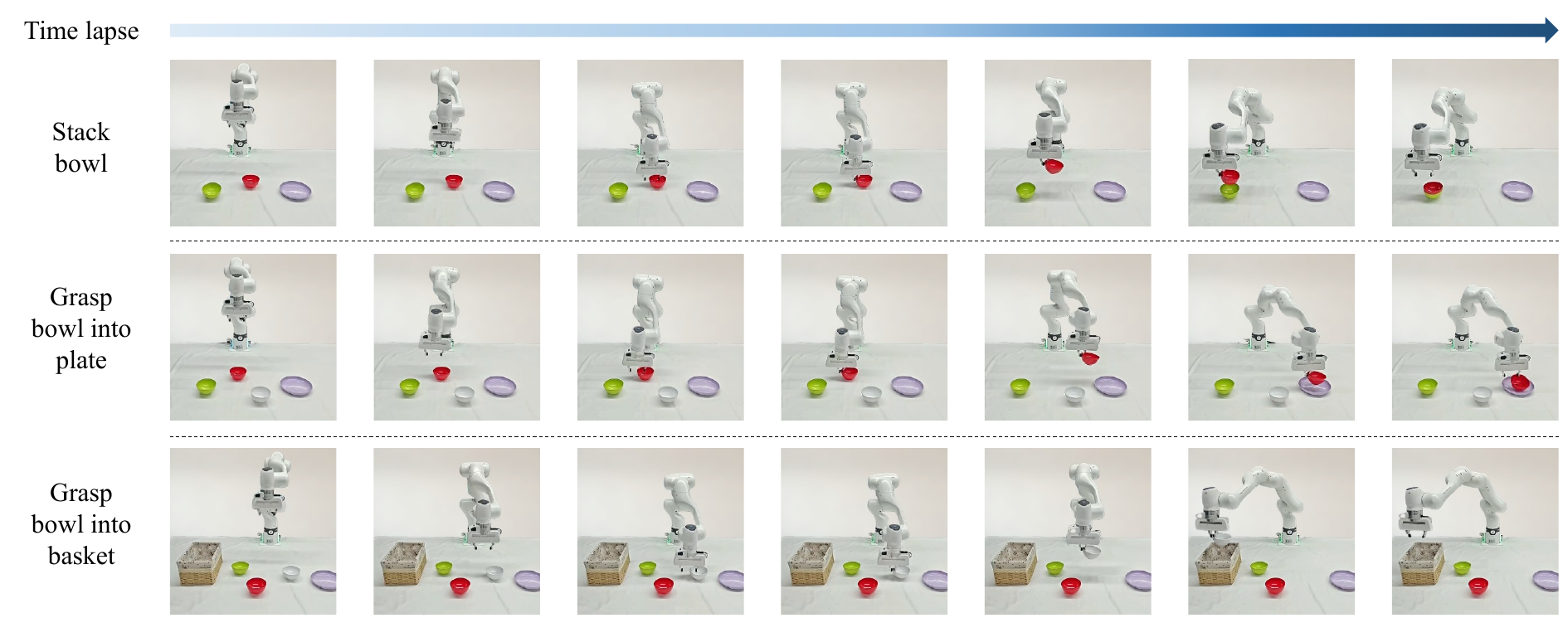}
    \caption{Real-world execution trajectories. Our specialist world model enhance the zero-shot generalization of the pre-trianed VLA model for diverse manipulation tasks.}
    \label{fig:real-execution}
\end{figure*}

\subsection{Ablation Studies}\label{sec:ablation}

\textbf{Module effectiveness}.
We perform ablation experiments to isolate the contributions of each component, as reported in Tab.~\ref{tab:ablation}.
When the MP module is removed, the model exhibits faster degradation during rollouts, resulting in an approximately 1.5\% drop in success rate.
Eliminating the TS-TTT module reduces generalization to unseen objects and environments, yielding a 3.5\% decrease.
Combining both modules leads to the best average success rate of 41.5\%, confirming that temporal memory and test-time adaptation play complementary roles.

\textbf{Variants of TS-TTT}. In the TS-TTT module, the temporal and spatial dimensions are combined for self-learning. Here, we explore different configurations to analyze the effectiveness of various combinations in Tab.~\ref{tab:TS-TTT}. First, the temporal, spatial and channel dimensions are conduct the self-learning individually. This setting yields suboptimal performance, indicating an implicit relationship between the spatial and temporal dimensions. Combining the channel dimension with either the spatial or temporal dimension results in worse performance, suggesting that channel-wise self-learning is unsuitable for fusion with other dimensions. Notably, fusing all three dimensions fails to converge, resulting in a `not-a-number' (NaN) loss.

\textbf{Cross-policy evaluation}.
We further evaluate our specialist world model in combination with different VLA models, including CogACT~\cite{li2024cogact} and $\pi_0$~\cite{black2024pi_0}.
As summarized in Tab.~\ref{tab:4}, the adapted world model significantly improves the performance of the pretrained CogACT by 41\%, demonstrating the strong generalizability of the world model on unseen scene. For the slightly better initial strategy $\pi_0$, our method also increases its average success rate by 27.0\%, further indicating its architecture-agnostic compatibility.




\subsection{Real-world Experiment}
To assess real-world transferability, we deploy the adapted world model on a Franka Research 3 \cite{franka_research_3} robotic arm (7 degrees of freedom: 6 end-effector + 1 gripper), as shown in Fig.~\ref{fig:setup}.
The environment consists of a table workspace with household objects of varying shapes and textures.
We collect 200 video clips of robot–environment interactions, covering grasping, placing, and pushing actions under natural lighting.
Following the same fine-tuning protocol as in simulation, the model is trained for 10,000 iterations using the real-world dataset. 
We design five physical tasks—pick-and-place, drawer closing, object stacking, cup pouring, and door pushing—to test robustness under visual noise and domain shift.
Each task is executed 20 times, and we report the average success rate. 

As shown in Fig.~\ref{fig:real-execution}, our specialist world model enhances the zero-shot generalization of the pretrained VLA model in unseen real-world environments for diverse tasks such as \emph{stack bowl}, \emph{grasp bowl into plate}, and \emph{grasp bowl into basket}. More specifically, AdaPower improves the average success rate of the pretrained CogACT by over 30\%. More detailed results and analyses are provided in the Supplementary Material.




\section{Conclusion}
We introduced \textbf{AdaPower}, a lightweight yet powerful framework for adapting a world foundation model to a specialist world model for action-conditioned predictive manipulation. AdaPower integrates a Temporal-Spatial Test-Time Training (TS-TTT) module, which enhances generalization via self-supervised adaptation across spatial-temporal dimensions, and a Memory Persistence (MP) module, which preserves long-horizon consistency through historical cross-attention. 
Deployed within a MPC framework alongside pre-trained VLAs, 
our approach enhances zero-shot generalization to unseen environments, achieving 30-40\% improvement in success rates across simulation and real-world settings.
While effective, our method introduces moderate test-time overhead, which may affect execution performance. Future work will address domain transfer robustness, real-time efficiency, and broader deployment in intelligence systems such as autonomous driving and multi-agent interaction.
{
    \small
    \bibliographystyle{ieeenat_fullname}
    \bibliography{main}
}


\end{document}